\documentclass[a4]{article}
\usepackage{graphicx}

\def\be{\begin{equation}}
\def\ee{\end{equation}}
\newcommand{\ff}[1]{{\bf #1}}

\def\lam{\lambda}
\def\x{\ff{x}}
\def\v{\ff{v}}

\begin{document}

\title{Swarm Intelligence: Past, Present and Future}

\author{Xin-She Yang$^1$, Suash Deb$^2$, Yu-Xin Zhao$^3$, 
 Simon Fong$^4$, Xingshi He$^5$ \\[10pt]
1)  School of Science and Technology, Middlesex University,
 London NW4 4BT, UK. \\
2) IT \& Educational Consultant, Ranchi, India. Also, Distinguished Professorial Associate,\\ Decision Sciences and Modelling Program,
Victoria University, Melbourne, Australia. \\
3) College of Automation,  Harbin Engineering University,  Harbin, China. \\
4) Department of Computer and Information Sciences,
University of Macau,  Macau, China.\\
5) College of Science,  Xi'an Polytechnic University,
No. 19 Jinhua South Road, Xi'an, China.
}

\date{}
\maketitle

\begin{abstract}
Many optimization problems in science and engineering are challenging to solve, and the current trend is to use swarm intelligence (SI) and SI-based algorithms to tackle such challenging problems. Some significant developments have been made in recent years, though there are still
many open problems in this area. This paper provides a short but timely analysis
about SI-based algorithms and their links with self-organization. Different characteristics
and properties are analyzed here from both mathematical and qualitative perspectives.
Future research directions are outlined and open questions are also highlighted.\\[10pt]

\noindent {\bf Citation Details:}
X.-S. Yang, S. Deb, Y.X. Zhao, S. Fong, X.S. He,
Swarm intelligence: past, present and future,
{\it Soft Computing}, (2017).
https://doi.org/10.1007/s00500-017-2810-5
\end{abstract}


\section{Introduction}

Many optimization problems in science and engineering applications
are highly complex and challenging, and thus require novel problem-solving approaches.
Traditional approaches tend to use problem-specific information such as the gradients of
the objective to guide the search for optimal solutions, and such approaches tend to be
highly sophisticated and specialized. They also have the disadvantage of getting
trapped in local optima, except for linear programming and convex optimization.
One of the current trends is to solve difficult optimization
problems in a quasi-heuristic way in combination with the successful characteristics of multi-agent systems. Such trend seems also to be true for solving problems in industry and business settings.
This new way of problem solving has resulted in a significant
development of new and novel swarm intelligence based algorithms.

In nature, many living organisms live in a community where there is
no centralized decision-making. In fact, the decision making among many
biological systems, especially social insects such as ants and bees, seems to occur
in a distributed, local manner. Individuals make decisions based on local information
and interactions with other agents and their environment. Such local interactions
seem to be responsible for the rise of social intelligence, and it can be hypothesized
that such complex interactions may directly or indirectly somehow contribute to the
emergence of intelligence in general. After all, changes tend to be some sort of responses and
adaptation to the changes of the organism's community and environment.
Groups of different organisms of the same species in nature have been found to be
successful in carrying out specific tasks, by means of a collective behaviour, namely
collective intelligence or swarm intelligence (SI) \cite{Fisher,Miller,Suro}.

It has also been observed in nature that different species can also co-evolve
and cooperate under the right conditions, especially when the resources are sparse.
Such swarm intelligence has inspired researchers to develop various ingenious ways for
solving challenging problems in optimization, machine learning and data mining \cite{BDorigo,Kennedy,YangBA,YangDeb}. Nature-inspired algorithms tend to be flexible,
easy to implement and sufficiently versatile to deal with different types of optimization
problems in practice. Such characteristics enable to solve problems that may be too challenging to solve using traditional algorithms.

Accompanying the emergence and success of nature-inspired algorithms, especially
the SI-based algorithms, there is a strong need to understand the mechanisms of algorithms in a
rigorous mathematical perspective. In contrast, the progress in theory lags behind.
Thus, it is often the case that we know how to use such algorithms and know they will generally work well, but we rarely know why they work under exactly what conditions. Consequently,
the use and applications of such metaheuristic algorithms are partially heuristic as well.

However, some promising progress starts to emerge in recent years concerning the analysis of
algorithms using Markov chain theory, dynamic systems, random walks
and stability analysis. They start to provide some insight into the intrinsic part of
algorithms. This paper will briefly review the   state-of-the-art developments concerning
swarm intelligence with a focus on both the present and future. We will also highlight
some key challenges and trends for future developments.
Therefore, the paper is organized as follows. Section 2 first briefly touches the concept of swarm intelligence and then Section 3 mainly focuses on
the present, and Section 4 looks at these algorithms from a theoretical perspective.
Section 5 will try to inspire the future research. Finally, the paper concludes
briefly in Section 6.

\section{Swarm Intelligence: A Critical Analysis}

Swarm intelligence can arise in multi-agent systems and it is not clear yet what mechanisms are responsible for the emergence of collective behavior in a swarm. Even so, swarm-intelligence-based algorithms have been developed and applied in a vast number of applications in optimization,
engineering, machine learning, image processing and data mining. Here, we review critically the essence of swarm intelligence and its link with self-organization.

\subsection{Swarm Intelligence}

The emergence of swarm intelligence (SI) is a complex process, and
it is not quite clear what mechanisms are required
to ensure the emergence of collective intelligence.
Inside a swarm, individual agents such as ants and bees in the complex system follow simple rules, act on local information, and there is no centralized control \cite{BDorigo,Miller}. Such rule-based interactions can lead to the emergence of self-organization, resulting in structures and characteristics at a higher system level. Loosely speaking, individuals in the system are not intelligent, but the overall system can behave intelligently, at least as can be considered
as some sort of collective intelligence. Such emerging self-organization can explain some key
swarming behaviour from ants to people \cite{Fisher,Suro}.

For such self-organization behaviour to emerge, it seems that there are certain
conditions that are necessary, and conditions such as feedbacks, stigmergy, multiple interactions, memory and environment setting are very important.
but it is still not clear about the exact role of such conditions and how seemingly self-organized structures can arise under such conditions. Though there are different attempts that try to understand the system behaviour, however, different studies in various subjects typically focus on one or a subset of these factors \cite{Corne,BDorigo,Fisher,Keller,Pars,Ting}.

Even though we may not fully understand the true mechanisms that lead to the self-organization and intelligent characteristics of a complex system, researchers have successfully developed optimization algorithms based on swarm intelligence. Examples of such algorithms
include particle swarm optimization (PSO) \cite{Kennedy}, ant colony optimization (ACO) \cite{BDorigo}, bat algorithm (BA) \cite{YangBA}, cuckoo search (CS) \cite{YangDeb},
flower pollination algorithm (FPA) \cite{Yang2014}, wolf search algorithm (WSA) \cite{Fong} and many others.

Before we discuss the links between swarm intelligence, self-organization and algorithms, let us analyze first the main characteristics of optimization algorithms.

\subsection{Algorithmic Characteristics }

Algorithms have always been an important part of computation \cite{Berlin,Chab}, but contemporary algorithms tend to be a combination of deterministic and stochastic components. Almost all nature-inspired algorithms use some aspects of swarm intelligence with stochastic components \cite{YangIEEE}. Since swarm intelligence-based algorithms are very diverse in terms of the sources of inspiration in nature and their formulations, there are different ways of analyzing and decomposing the essential components of these algorithms. For example, we can look at algorithms by focusing on the key characteristics and their properties from a perspective of self-organizing systems.
This is a higher-level analysis that does not depend on the details of the mathematical formulations or algorithmic steps, which allows us to focus on the functionalities and the main search behaviour of algorithms.

First, all SI-based algorithms use a population of multiple agents and each agent is represented by
a solution vector $\x_i$, and each vector can be considered as a state of the system. An algorithmic system is typically initialized by setting the population as the random sampling of the search space. Then, the update of this population is realized by moving the agents in a quasi-deterministic manner to be referred to as its `algorithmic dynamics'. This algorithm dynamics determines how the
system evolves, according to a set of equations (such as those used in particle swarm optimization) or a predefined procedure (such as those used in genetic algorithms). Randomness is often used in  SI-based algorithms to act as a perturbation force to drive the system from equilibrium and potentially jump out of local valleys in the objective landscape.

In addition, a selection mechanism is needed to select the best solution (or the fittest solutions) in the population so as to allow the fittest to pass onto the next generation. This means that
some states/solutions are preferably selected. Such selection, together with the evolution of population, often enables the population in the search process to converge to a set of solutions
(often the optimal set), and consequently some convergent states or solutions may emerge as iterations continue.

For example, let us consider both particle swarm optimization (PSO) and firefly algorithm (FA)
to be introduced later. Both algorithms have randomization by using random numbers. Though their use of two random numbers is different, both can provide some form of stochastic properties in generating new solutions so that new solutions can be different and sometimes sufficiently distant from existing solutions.
This means that they essentially provide the ability for the algorithms to escape local optima without being trapped.

However, there are significant differences between PSO and FA.
Firstly, PSO uses the best solution found so far $g_*$, while FA does not use $g_*$. Secondly,
PSO is a linear system, while FA is nonlinear in terms of updating equations. Thirdly, the attraction mechanism in FA allows the swarm to subdivide into multiple small subswarms, which
enables FA to solve multimodal problems more effectively. On the other hand, PSO cannot subdivide
the swarm. In addition, PSO has the drawbacks of using velocity, while FA does not use velocity.
Thus, these differences in algorithmic dynamics will lead to significantly different characteristics, performance and efficiency of algorithms. In fact, studies show that FA can have a higher convergence rate in most applications \cite{Fister,Tila}.

In all algorithms, iterations are used to provide the evolution of solutions towards some selected solutions in terms of a pseudo-time iteration counter. At the initial stages of such iterative evolution, solutions tend to have much higher diversity
as solutions are usually different and often uniformly distributed randomly in the search space. As the evolution continues, solutions become more similar to each other by some selection mechanism based on the fitness landscape. Selection acts as a driving force for evolution. Good solutions are selected according to their fitness, often the objective values, which exerts a selection pressure for the multi-agent populations to adapt and react to the changes in the objective landscape and can thus drive the system to converge towards some specific, selected states or solutions.

These key characteristics and properties as well as their role can be summarized
in Table \ref{table-char}.
It is worth pointing out that this is only one way of looking at the algorithms
and the emphasis here
is purely for the convenience of comparing with the mechanisms for self-organization to be discussed in the next subsection.
\begin{table}
\begin{center}
\caption{Main characteristics of an algorithm based on swarm intelligence. \label{table-char} }
\begin{tabular}{|l|l|l|}
\hline
Algorithmic Components & Characteristics & Role/Properties \\
\hline \hline
Multi-agents & Population & Diversity and sampling \\
\hline
 Randomization & Perturbations &  Escape local optima \\
\hline
 Selection & Driving force & Organization and convergence \\
\hline
 Algorithmic equations & Iterative evolution & Evolution of solutions \\
\hline
\end{tabular}
\end{center}
\end{table}

Obviously, there are other ways to look at algorithms \cite{Blum,Corne,YangSTA}. For example, the use of exploration
and exploitation is another good way to analyze the behaviour of algorithms \cite{Blum}.
In addition, mathematical analysis
can provide insight from a theoretical point of view \cite{YangSTA}. In general, different ways of looking at algorithms
can lead to different insights and thus understand the algorithms from different perspectives.

\subsection{Algorithms as Self-Organization}

A complex system may be able to self-organize under the right conditions. Loosely speaking,
when the size of the system is sufficiently large, it will lead to a sufficiently
high number of degrees of freedom or possible states $S$.
At the same time, there should be a sufficiently long time for the system to evolve
from noise and far from equilibrium states \cite{Ashby}.

Another important factor is that a proper selection mechanism must be in place to ensure that self-organization is
possible. In other words, the primary conditions for self-organization to evolve in a complex system
can be summarized as follows \cite{Ashby,Keller}:
\begin{itemize}
\item[$\bullet$] The size of the complex system is sufficiently large with
a higher number of degrees of freedom or states.

\item[$\bullet$] Enough diversity exists in the system in terms of
perturbations, noise, edge of chaos, or far from the equilibrium.

\item[$\bullet$] The system is given enough time to evolve.

\item[$\bullet$] There is a selection  mechanism (or an unchanging law)
in the system to select certain states.

\end{itemize}

If we loosely represent the above conditions mathematically, we can say that a
system with multiple states $S_i$ can evolve towards the self-organized states $S_*$,
driven by a driving mechanism $M(t,p)$ with a set of parameters $p$ that may vary with time $t$,
which can be written schematically as
\be S_i \stackrel{M(t,p)}{\Longrightarrow} S_*. \ee
If we look at an algorithm from the perspective of self-organization,
we can indeed consider  an algorithm  as a self-organization system,
starting from many possible states $\x_i$ (solutions) and tries
to converge to the optimal solution/state $\x_*$, driven by the selection mechanism in an
algorithm $A(p, t)$ with a set of parameter $p$, evolving with time pseudo-time $t$.
This can be represented in the following schematic format:
\be f(\x_i) \stackrel{A(p, t)}{\Longrightarrow} f_{\min}(\x_*) \; \textrm{ or } \; f_{\max}(\x_*). \ee

Now if we look both algorithms and self-organization systems more closely, we can identify the main role and properties of an algorithm and compare them with the conditions for
self-organization, we can summarize them in Table~\ref{table-alg}.
\begin{table}
\begin{center}
\caption{Similarities between self-organization and an algorithm. \label{table-alg} }
\begin{tabular}{|l|l||l|l|}
\hline
Self-organization & Features & Algorithm & Properties \\
\hline \hline
Multiple states & High complexity & Population & Diversity and sampling \\
\hline
Noise, perturbations & Diversity & Randomization & Escape local optima \\
\hline
Selection mechanism & Structure & Selection & Convergence \\
\hline
Re-organization & State changes & Evolution & Evolution of solutions \\
\hline
\end{tabular}
\end{center}
\end{table}

Despite these striking similarities, however, there are some significant differences
between a self-organizing system and an algorithm. First, for self-organization,
the exact avenues to the self-organized states may not be clear. But for an algorithm, the way that makes an algorithm converge is crucial. Second, time is not an important factor for self-organization, while the rate of convergence is paramount for an algorithm because the minimum computational cost is needed in practice so as to quickly reach either truly global optimality or suboptimal solutions. Finally, the structure can be important for a self-organized system, while the converged solution
vectors (rather than their structure) is more important for solving an optimization problem.

It is worth pointing out these similarities of self-organization to algorithms are applicable
for almost all stochastic algorithms, including the
classic stochastic algorithms such as genetic algorithm \cite{Gold}.
In addition, even we can consider an algorithm as a self-organized system,
this does not mean that we can always make an algorithm efficiently. This is partly
because the exact behaviour is influenced by both the interactions of algorithmic
components and algorithm-dependent parameters, but these details are not clearly
understood yet for most algorithms.

In the next section, we will outline the state-of-the-art developments
of SI-based algorithms before we proceed to do some in-depth mathematical analysis afterwards.

\section{The Present Developments}

In the current literature, there are many algorithms that use the concept of swarm intelligence (SI), and the number of SI-based algorithms is increasingly almost monthly. Therefore,
it is not possible to introduce and analyze these algorithms in a very short paper as such. Therefore, our emphasis will be on the brief analysis of a few selected algorithms as representatives so as to highlight the main points. Now we first introduce briefly a few algorithms and we then categorize them in terms of their characteristics and algorithmic dynamics
and links to self-organization.

\subsection{Algorithms Based on Swarm Intelligence}

Though both ant colony optimization (ACO) \cite{BDorigo} and particle swarm optimization (PSO) \cite{Kennedy} are primary examples of SI-based algorithms. However, ACO can be considered as a mixture of descriptive procedure and equations, while PSO is mainly based on dynamic equations. For this reason, we focus first on the PSO here.

 For the ease of discussing particle swarm optimization (PSO), developed by Kennedy and Eberhart \cite{Kennedy}, we use $\x_i$ and $\v_i$ to denote the position (solution) and velocity, respectively,  of a particle or agent $i$. The main iteratively updating equations for PSO are
\be \ff{v}_i^{t+1}= \ff{v}_i^t  + \alpha \ff{\epsilon}_1
[\ff{g}^*-\x_i^t] + \beta \ff{\epsilon}_2 [\x_i^*-\x_i^t],
\label{pso-speed-100}
\ee
\be \x_i^{t+1}=\x_i^t+\ff{v}_i^{t+1}, \label{pso-speed-140} \ee
where $\ff{\epsilon}_1$ and $\ff{\epsilon}_2$ are two uniformly distributed random vectors in [0,1].
Both $\alpha$ and $\beta$ are so-called learning parameters. This algorithm uses the
current global best solution $\ff{g}^*$ found so far as well as the individual best
$\x_i^*$. PSO has been applied in many areas in science and engineering \cite{Banks,Solei},
and it has also been extended to solve multiobjective optimization problems \cite{Reyes}.
For comprehensive reviews, please refer to \cite{Banks,Khare}.

It is clearly seen that the above algorithmic equations are linear in the sense that both equations
only depend on $\x_i$ and $\v_i$ linearly. Selection is carried out by the attractor or converged state  $\ff{g}^*$, which is also evolving. Randomization is done by two uniformly distributed random numbers.


Bat algorithm (BA) is another example of SI-based algorithms. BA was developed by Yang
and BA mainly uses frequency-tuning and some characteristics of echolocation of microbats \cite{YangBA}. The main algorithmic equations for BA are
\be f_i =f_{\min} + (f_{\max}-f_{\min}) \beta, \label{f-equ-150} \ee
\be \ff{v}_i^{t} = \ff{v}_i^{t-1} +  (\x_i^{t-1} - \x_*) f_i , \ee
\be \x_i^{t}=\x_i^{t-1} + \ff{v}_i^t,  \label{f-equ-250} \ee
where $\beta \in [0,1]$ is a random vector drawn from a uniform distribution.
$f_{\min}$ and $f_{\max}$ are the frequency-tuning range. These equations are also associated with the pulse emission rate $r$ and loudness $A$ that can be switched on or off by comparing with a uniformly distributed random number $\varepsilon$. For each bat $i$, we can use
\be r_i^{t+1}=r_i^{(0)} (1-e^{-\gamma t}), \quad A_i^{t+1} =\alpha A_i^t, \ee
where $0<\alpha<1$ and $\gamma>0$ are two parameters to control the variations of $r$ and $A$.

 The bat algorithm has been extended to multiobjective optimization and hybrid versions as well as chaotic bat algorithm with many applications \cite{Rodrig2,Osaba2,Kashi,YangHe,Gandomi}.

The algorithmic equations in BA are also linear in the sense that the equations depend on
$\x_i$ and $\v_i$ linearly.  However, the control of exploration and exploitation is carried out
by the variations of loudness $A(t)$ from a high value to a lower value, while the pulse emission rate is increased nonlinearly from a lower value to a higher value. Selection is done by
the current best solution $\x_*$, which acts a similar role as the $\ff{g}^*$ in PSO. Randomization is done by a uniformly distributed number $\beta$ for frequency tuning. As a result, BA can have a faster convergence rate.


Firefly algorithm (FA) developed by Yang is an algorithm inspired by the swarming behaviour of tropical fireflies. FA uses a nonlinear system by combing the exponential decay of light absorption and inverse-square law of light variation with distance. The main equation in FA is a single nonlinear equation in the following form:
\be  \x_i^{t+1} =\x_i^t + \beta_0 e^{-\gamma r^2_{ij} }
(\x_j^t-\x_i^t) + \alpha \; \ff{\epsilon}_i^t, \label{FA-equ} \ee
where $\alpha$ is a scaling factor controlling the step sizes, while $\gamma$ is
a scale-dependent parameter controlling the visibility of the fireflies (and thus
search modes). In addition, $\beta_0$ is the attractiveness constant.
Firefly algorithm has been applied to many applications \cite{Carbas,Darwish,Fister,Fister2,Galvez,Marich2,Osaba,Senth,Tila,Zhao} and there are many different variants such as neighborhood firefly algorithm \cite{Wang} and the quantum-based hybrid \cite{Zoua}.

Since FA is a nonlinear system, it has the ability to automatically subdivide the whole swarm into
multiple subswarms due to the fact that short-distance attraction is stronger than long-distance attraction. Each subswarm can potentially swarm around a local mode, and among all the modes, there is always a globally optimal solution.
Therefore, it is suitable for multimodal optimization problems. There is no explicit use of the best solution, thus selection is through the comparison of relative brightness according to the rule of `beauty is in the eye of the beholder'. Randomization is explicitly done by a perturbation term in the equation.
As pointed out earlier, FA has some significant differences from the PSO. FA is nonlinear, while the PSO is linear. FA has an ability of multi-swarming, while the PSO cannot. In addition, the PSO uses velocities
(and thus some drawbacks), while FA does not use velocities. Furthermore, FA have a scaling control by using $\gamma$, while the PSO has no scaling control. All these differences enable FA to search more effectively for multimodal objective landscapes.


Cuckoo search (CS), developed by Yang and Deb, is another nonlinear system.
CS is a primary example of intriguing brooding parasitism of some cuckoo species, and
it uses a balanced combination of both local and global search capabilities,
controlled by a switching probability $p_a$. One equation is mainly local and can be written as
\be \x_i^{t+1}=\x_i^t +\alpha s \otimes H(p_a-\epsilon) \otimes (\x_j^t-\x_k^t), \label{CS-equ1} \ee
where $\x_j^t$ and $\x_k^t$ are two different solutions selected randomly by random permutation,
$H(u)$ is a Heaviside function, $\epsilon$ is a random number drawn from a uniform distribution, and $s$ is the step size.

The other equation is mainly global and can be expressed as
\be \x_i^{t+1}=\x_i^t+\alpha L(s,\lam), \label{CS-equ2} \ee
where the L\'evy flights are simulated by
\be L(s,\lam) \sim \frac{\lam \Gamma(\lam) \sin (\pi \lam/2)}{\pi}
\Big(\frac{1}{s^{1+\lam}}\Big), \quad (s \gg 0). \ee
Here $\alpha>0$ is the step size scaling factor.
Cuckoo search has become powerful in solving many problems such as software testing \cite{Prav}, scheduling \cite{Marich}, cyber-physical systems \cite{Cui} and others \cite{YangDeb,Zine}.

As we can see from the above equations, CS is a nonlinear system due to the Heaviside function and
switch probability. Selection is not via the explicit use of global best $\ff{g}^*$, but selection is done by ranking and elitism where the current best is passed onto the next generation.
Randomization is carried out more effectively using L\'evy flights where a fraction of steps are
larger than those used in Gaussian. Thus, the search is heavy-tailed \cite{Reyn}. In addition, as the L\'evy flights can be approximated by a power-law type of distribution, the search steps are also scale-free. In fact, it is observed in the simulations that CS is indeed scale-free and have a
fractal-like search structure. Consequently, CS can be very effective for nonlinear optimization problems and multiobjective optimization \cite{Gandomi2,Ouaa,Ma,YangDeb,Zine}.

Flower pollination algorithm (FPA) is inspired by the pollination characteristics of flowering plants \cite{Yang2014}, and FPA mimics the biotic and abiotic pollination characteristics as well as the flower constancy as a co-evolution between certain flower species and pollinators such as insects and animals. The main algorithmic equations are
\be \x_i^{t+1}=\x_i^t + \gamma L(\lam) (\ff{g}_* - \x_i^t), \ee
and
\be \x_i^{t+1} =\x_i^t + U \; (\x_j^t - \x_k^t), \ee
where $\gamma$ is a scaling parameter, $L(\lam)$ is the random number vector drawn from a L\'evy distribution governed by the exponent $\lam$.
Here $\ff{g}_*$ is the best solution found so far, which acts as a selection mechanism. In addition, $U$ is a uniformly distributed random number. Furthermore, $\x_j^t$ and $\x_k^t$ are solutions representing pollen from different flower patches.

FPA is a quasi-linear system because the equations are linear in terms of $\x_i$, but the random switching between two branches of search moves introduces some weak nonlinearity. Selection uses
the current best explicitly, while randomization is carried out via three components: L\'evy flights, a uniform distribution and a switch probability. Thus, FPA can have a higher explorative ability while remaining a strong exploitation ability. In fact, it has recently been proved that FPA can have guaranteed global convergence under the right conditions \cite{He2017}. FPA has been applied to many applications with an expanding literature \cite{Alam,Bekdas,Rodrig,ZhouY}.

Obviously, there are other SI algorithms, including the wolf search algorithm (WSA) \cite{Fong}, ant colony optimization (ACO), artificial bee colony and others,
but we do not have space to discuss them in this paper.

In addition, some algorithms such as differential evolution (DE) are also very efficient \cite{Storn}, but they may not be classified as SI-based algorithms, and thus we will not discuss them either.
Instead, our focus will be on the discussion and analyses of the above algorithms.

It is worth pointing out that though there are many algorithms in the literature, there is no single algorithm that can be most efficient to solve all types of problems as dictated by the no-free-lunch theorems \cite{Wolpert}. However, under the right conditions such as co-evolution, certain algorithms can be more effective \cite{Wolpert2}. As our purpose here is not to search for best algorithms, our main focus is to gain more insights into different types of algorithms. In the rest of section, we will discuss the main characteristics of SI-based algorithms.

\subsection{Characteristics of SI Algorithms}

Based on the above brief descriptions of some SI-based algorithms, we can now
analyze them in terms of their main characteristics such as randomization techniques,
selection mechanism, the potential mechanism driving the evolution, as summarized in Table \ref{table-sum}. Here, we include genetic algorithms (GA) for the purpose of comparison \cite{Gold}.

\begin{table}
\begin{center}
\caption{SI Algorithms and their relevant characteristics. \label{table-sum} }
\begin{tabular}{|l||l|l|l|}
\hline
Algorithm & Randomization & Selection & Evolution Mechanism \\
\hline \hline
GA & Uniform & Elitism  & Survival of the fittest \\
\hline
PSO & Uniform & $\ff{g}^*, \x_i^*$ & swarming towards $\ff{g}^*$ \\
\hline
BA & Uniform & $\x_*$ & swarming towards $\x_*$ \\
\hline
FA & Gaussian & Brightest & Attraction \\
\hline
CS & L\'evy flights & Best & Similarity/Elitism \\
\hline
ACO & Probabilistic & Pheromone & Pheromone variations \\
\hline
FPA & Uniform \& L\'evy & $\ff{g}^*$ & constancy and similarity \\
\hline
\end{tabular}
\end{center}
\end{table}

From this table, we can see that almost all algorithms use some sort of best solutions
such as the centre of the swarm. Some algorithms such as PSO, BA and FPA use the current
global best solution explicitly in their formulations, while others such as FA, CS and ACO
use it in an implicit way.  One of the advantages of explicit use of $\ff{g}^*$  is that
it provides a direct driving force in the governing equations, and thus it may be able
to speed up the convergence. However, if this driving force is too strong, it may lead
to premature convergence as it can often be observed in PSO. On the other hand,
the advantage of implicit use (either by ranking or post-processing) can lead to
a higher probability of finding the true global best solution, thus potentially
avoid some form of premature convergence. However, it may slow down the search process
due to a weaker driving force for evolution.

Another advantage without the direct use of $\ff{g}^*$ is
that multiswarms can occur as in the case of FA due to its nonlinear attraction among different fireflies. In the standard firefly algorithm, as the short-distance attraction is stronger than long-distance attraction, the whole population can automatically subdivide into many sub-swarms, and each sub-swarm can potentially swarm around a local mode, and among all the modes, there is certainly the global best solution.  All this makes it natural for FA to deal with mutlimodal optimization problems effectively.

It is worth pointing out that the above analysis is just one way to analyze SI-based algorithms from a higher-level but qualitative perspective. Another way of analyses is to use rigorous mathematical theories, which will be the focus of next section.

\section{Mathematical Framework for Algorithms}

Mathematical frameworks for analyzing algorithms can be dynamic systems, fixed-point theory, Markov chain theory,
self-organization, filtering and others. Though our long-term intention is to build a solid mathematical framework to analyze algorithms, however, it is not possible to achieve such a huge task in a single paper. Instead, we would like to highlight a few key points so as to inspire the research community to carry out more research in this area.

\subsection{Fixed-Point Theory}
Traditional numerical analysis tends to focus on the iterative nature of an algorithm $A(\x_t)$
and see how the solution $\x_t$ evolves as a pseudo-time iteration counter $t$. From the well-known
Newton's method for finding the roots of a nonlinear function $f(\x)$
\be \x_{t+1} =\x_t -\frac{f(\x_t)}{\nabla f(\x_t)}, \ee
we know that when the solution sequence converges, we have
\be \lim_{t \rightarrow \infty} \x_{t} = \x_*, \ee
where $\x_*$ is the final converged solution which is essentially the fixed point. The general theory is the fixed-point theory which dictates how an iterative formula may evolve and lead to a fixed point in the search space \cite{Suli}.

For a population of solutions in any of SI-based algorithms, the population can interact with each other and may lead to potentially multiple fixed points, depending on the algorithm dynamics of each  algorithm. It can be expected that $\ff{g}^*$ acts as a fixed point in PSO, while there are multiple fixed points in the firefly algorithm. In fact, we can hypothesize that there is a single fixed point in BA, PSO, simulated annealing, FPA and bee algorithm, while multiple fixed points can exist in FA, CS, ACO and genetic algorithms if the conditions are right. However, it is not clear yet what these conditions should be, and such conditions may also be problem dependent. More research is highly needed in this area.

\subsection{Dynamic System}
The first analysis of PSO using a dynamic system theory was carried out by  Clerc and Kennedy \cite{Clerc}, and they linked the governing equations of PSO with the dynamic behaviour of particles
under different parameter settings. In fact, their analysis suggested that the PSO system
is governed by the eigenvalues of a system matrix
\be \lam_{1,2} =1-\frac{\gamma}{2} \pm \frac{\sqrt{\gamma^2-4 \gamma}}{2}, \ee which leads a bifurcation at $\gamma=\alpha+\beta=4$. Such analysis can indeed provide some insight into the
working mechanism and main characteristics, however, they do not provide a full picture of the
system due to the assumptions and simplifications used in the analysis.

Though in principle we can use the similar method to analyze other algorithms, however, it becomes
difficult to extend to a generalized system. For example, in FA, CS and ACO, the nonlinearity makes it difficult to figure out the eigenvalues because the matrix will depend on the current solution,
randomization and other factors. In addition, nonlinearity in FA also means that the characteristics can be much richer than simple linear dynamics such as PSO. Thus, this method may become intractable in general and it may not be very useful to gain any insight into these algorithms.

\subsection{Markov Chain Theory}

From the probability perspective, the solutions generated by an algorithm is a statistical sampling method such as Monte Carlo \cite{Fishman}. In the more general sense, the solution set generated by an algorithm essentially form a system of Markov chains. A Markov chain is a chain whose next state will depend only on the current state and the transition probability. In this sense, Markov chain theory can provide a generalized framework for analyzing SI-based algorithms. In fact,
a simple analysis of genetic algorithms using Markov chain theory was carried out by Suzuki \cite{Suzuki}, and a discrete-time Markov chain approach has been used to prove that the flower pollination algorithm can have guaranteed global convergence \cite{He2017}.

On the other hand, a generalized approach has been designed using a
Markov chain  for global optimization \cite{Ghate}; however, this approach may converge slower than SI-based algorithms. This methodology can provide a quite general framework for optimization.

It is worth pointing out that Markov chain theory can be rigorous, enabling to provide some significant insight into the algorithms. In theory, the largest eigenvalue of a proper Markov chain is one, while the second largest eigenvalue $\lam_2$ of the transition probability matrix essentially controls the rate of convergence of the Markov chain. But in practice it is very challenging to find this eigenvalue. Even some estimates can be difficult. Therefore, the information and insight we can obtain is limited in practice, which may also limit its practical use.

\subsection{Self-Organization}

As we have seen earlier, an algorithm can be considered as a self-organizing system where multiple agents sample the search space, driven by a selection mechanism, evolving according a predefined procedure or a set of algorithmic equations. The iterative evolution will usually lead to a converged set of solutions that may correspond to the optimal solutions to the problem under consideration. There are similarities and differences between algorithmic evolution and self-organization as summarized in Table 2. However, such comparison and perspective only provide the qualitative nature of the algorithm. Though the insightful can be at higher level, it lacks crucial details about how the self-organized states emerge, under what conditions and how quickly such converged states can be reached. Unless new theory about self-organization emerges, the information we can gain is mainly qualitative. Key information and properties may need to obtain by other means.

\subsection{Other Approaches}
Sometimes, it may not be easy to put some studies into a fixed category, but their results can be equally useful \cite{Villa}. For example, Zaharie carried out a variance analysis of population and the effect of crossover in differential evolution \cite{Zaha}. The variance provides some information about the diversity of the population during the iterations.

\subsection{Multidisciplinary Approach}

From the above discussion, it seems that one approach can give only a part of the full information and insight. Different approaches and perspectives can provide different insights, potentially
complementary to each other. Therefore, to truly understand a complex algorithm, it may be useful to use all different approaches so as to build a fuller picture about the algorithm. It can be expected that a multidisciplinary framework can be formulated to analyze algorithms comprehensively.

\section{Trends and Future Challenges}

The above analyses and discussions about SI-based algorithms have laid the foundation for us
to turn our attention to the possible future developments.
Obviously, it is not possible to predict what future research would be, but we hope to
inspire more studies in the important directions concerning swarm intelligence and their
applications in optimization, machine learning and data mining. Therefore, we would like to
highlight some of key challenges in this area.

\subsection{Some Key Challenges}

Many challenging issues exist concerning swarm intelligence, and it is not our intention
to address every aspect of these challenges. As an example, the emergence
of intelligent behaviour among a complex swarm is still poorly understood, which will not be
addressed here.  Therefore, we can only focus on a small but key set of challenging issues as outlined below.

\begin{itemize}
\item \emph{Parameter tuning and control}: All algorithms have algorithm-related parameters and
some algorithms have more parameters than others. In general, the setting of these parameters can affect the algorithm significantly, though some parameters may
have a weak influence, while others may have a strong influence. In theory, these parameters should be tuned so as to maximize the performance of the algorithm, however, such parameter tuning is not a trivial task \cite{Eiben}. Even with well-tuned parameters, there is no strong reason that they should remain fixed. It may be advantageous to use varying parameters during iterations and the proper variations of parameters are called parameter control. Both parameter tuning and
control can be considered as a high level of optimization; that is the optimization of optimization algorithms.

Currently, most tuning approaches are done using parametric studies, while parameter control uses stochastic adaptivity where certain parameters are allowed to vary randomly within a predefined range. Ideally, parameter tuning and control can be done automatically, such as the self-tuning framework by Yang et al. \cite{YangSTA}. However, the computational costs may be still high. Therefore, there is a strong need to find an effective way to tune parameters both  automatically and adaptively.

\item \emph{Optimal exploration and exploitation}: An efficient algorithm should be able to balance the exploration of the search space and the exploitation of the landscape information. Exploration can increase the diversity and thus increase the probability of finding the global optimal solutions, while exploitation uses local information to enhance the search process.
    However, too much exploration and too little exploitation will slow down the search process, while too much exploitation and too little exploration will lead to premature convergence. The optimal balance can be difficult to find, and empirical observations suggest that such balance may be also problem-specific. How to achieve such a balance is still
an open problem, though it is possible to produce a better balance under certain conditions.

\item \emph{Large-scale problems and algorithm scalability}: The current literature seems to suggest that SI-based algorithms can be effective in solving various design problems, and there is some indication that they can even solve highly complex NP-hard problems, but the case studies in the current literature have most about optimization problems with the number of variables ranging from a few to a few hundred. Compared to real-world applications, the dimensionality tested is relatively low. However, it is not clear if these algorithms can be directly applied to large-scale problems. The true scalability is yet to be tested. It is highly needed to test problems with the number of variables more than a thousand or even much higher.

\item \emph{Mathematical Framework}: As we discussed earlier, there are different ways of looking at SI-based algorithms and analyzing them from different perspectives such as stability, dynamic systems, Markov chain theory and self-organization. However, there is no unified framework yet that can provide a fuller picture of an algorithm, concerning convergence, rate of convergence, stability, ergodicity, repeatability and scalability. It is highly likely that any unified framework for theoretical analysis is a multidisciplinary approach, looking at algorithms from all angles and perspectives.

\item \emph{Rate of convergence and control}: From both theoretical and practical perspectives, the rate of convergence is extremely important. After all, we want the best solution to a problem quickly with the minimum computational costs. Even though we can understand largely the characteristics of many algorithms, this does not mean that we can control their behaviour, especially the rate of convergence in practice. Loosely speaking, the rate of convergence can depend on many factors such as the intrinsic components, structure, parameter values and initial configuration of an algorithm, and such dependence can be
    complex, indirect and nonlinear. Even in the case it may be possible to figure out the rate of convergence, it may be difficult to control it so as to maximize the search efficiency. It can be expected that such control can be interlinked with parameter tuning and control.

\end{itemize}

\subsection{Recommendations for Future Research}

With the key challenges we just outlined, it is highly recommended to carry out further research to
address such challenges. Therefore, research priority should be given to the following areas:

\begin{itemize}

\item \emph{Theoretical framework}: Due to its importance in understanding how algorithms work,  theoretical analysis should be among one of the top priorities in the near future.
    Theoretical analysis can gain more insight into algorithms that allow us to identify the best
    types of problems to solve and to tune or control the parameters more effectively. This may also allow us to potentially design better and more effective algorithms and tools.

\item \emph{Hybridization}: Though some algorithms are very effective in solving certain types or even a wider range of problems, studies suggest that hybridization can be powerful by combining the advantages of different algorithms \cite{Ting}. In fact, hybrid algorithms have been attempted for many years, but the hybridization process is still a bit trial and error.
    It is not clear yet how to combine different algorithms so as to produce a better hybrid?

\item \emph{Self-tuning and self-adaptive algorithms}: As we mentioned earlier, the tuning of algorithm-dependent parameters is a challenging task. The control of parameters
    is also a difficult task.
    Ideally, a truly useful algorithm should be able to self-tune and self-adapt to suit for different types of problems \cite{YangSTA}. The main unanswered questions are: what is the best way for algorithms to be self-tuning and self-adaptive? For a given set of algorithms, how to adapt them to new problems without any prior knowledge?

\item \emph{Diverse applications}: The usefulness of algorithms is the ability to solve a wide range of problems, especially large-scale, real-world applications. After all, there are  many optimization problems that need to be solved in all areas of science, engineering, industry and business applications.

\item \emph{Intelligent tools}: Several decades of intensive research in algorithms and optimization have enable researchers to design better and more effective tools. However,
    no one can claim that they have produced truly intelligent tools that can solve problems automatically, quickly and intelligently. In fact, there are so many related questions concerning this issue. For example, what do we mean by `intelligent algorithm'? Can algorithms really be intelligent? Questions like these can be endless, but we may at least wish to know what the minimum components are so as to make an algorithm sufficiently intelligent?

\end{itemize}

Obviously, there are other important directions and active research topics concerning swarm intelligence, optimization and machine learning. One important topic is to use a good combination of new algorithms with traditional techniques because traditional techniques have been well established and tested, and they are among the most useful ones to a specific class of problems. New methods will be most needed when traditional methods do not work well. In addition, even algorithms are efficient, the proper implementation and parallelization can make algorithms even more useful in practice.

\section{Conclusions}

Swarm intelligence is an interesting and important area, and swarm intelligence based algorithms have permeated into almost all areas of sciences and engineering. Accompanying their success and popularity, there are some key issues to be addressed.  In this paper, we have
first reviewed the essence of swarm intelligence, and then linked algorithms and swarm intelligence to self-organization of complex systems. Then, we highlighted some SI-based algorithms and subsequently analyzed their main components, characteristics and properties, followed by a more theoretical approach using fixed-point theory, dynamic system and Markov chain theory. Finally, we have also outlined some key challenges and provide some recommendations for addressing such issues. It is authors' hope that more research can be inspired, concerning swarm intelligence and nature-inspired computation
 so as to solve a diverse range of optimization problems in real-world applications.


\end{document}